\documentclass[sigconf]{acmart}

\pdfoutput=1
\renewcommand\footnotetextcopyrightpermission[1]{}
\setcopyright{none}

\usepackage{booktabs}
\usepackage{todonotes}
\usepackage{caption}
\usepackage{subcaption}
\usepackage[normalem]{ulem}

\newcommand{\XXX}{{\sc EBAnO}} 
\newcommand{\XXXdef}{{\sc E}xplaining {\sc B}l{\sc A}ck-box m{\sc O}del} 

\newcommand{\IndexRI}{{\sc IR} } 
\newcommand{\IndexRIDef}{{\sc I}nfluence {\sc R}elation } 

\newcommand{\IndexB}{{\sc IRP} }
\newcommand{\IndexBDef}{{\sc I}nfluence {\sc R}elation {\sc P}recision }

\begin{document}
\title{What's in the box? Explaining the black-box model through an evaluation of its interpretable features}

%\\\\\\\\\\\\\\\\\ Authors \\\\\\\\\\\\\\\
\author{Francesco Ventura}
\affiliation{%
  \institution{Politecnico di Torino}
  \streetaddress{Corso Duca degli Abruzzi, 24}
  \city{Torino}
  \state{Italy}
  \postcode{10129}
}
\email{francesco.ventura@polito.it}

\author{Tania Cerquitelli}
%\authornote{This author is the   one who did all the really hard work.}
\affiliation{%
  \institution{Politecnico di Torino}
  \streetaddress{Corso Duca degli Abruzzi, 24}
  \city{Torino}
  \state{Italy}
  \postcode{10129}
}
\email{tania.cerquitelli@polito.it}

\begin{abstract}
Algorithms are powerful and necessary tools behind a large part of the information we use every day. However, they may introduce new sources of bias, discrimination and other unfair practices that affect people who are unaware of it. Greater algorithm transparency is indispensable to provide more credible and reliable services. Moreover, requiring developers to design transparent algorithm-driven applications allows them to keep the model accessible and human understandable, increasing the trust of end users.
In this paper we present \XXX, a new engine able to produce prediction-local explanations for a black-box model exploiting interpretable feature perturbations. \XXX\ exploits the hypercolumns representation together with the cluster analysis to identify a set of interpretable features of images. Furthermore two indices have been proposed to measure the influence of input features on the final prediction made by a CNN model. \XXX\ has been preliminary tested on a set of heterogeneous images. The results highlight the effectiveness of \XXX\ in explaining the CNN classification through the evaluation of interpretable features influence.
\end{abstract}


\maketitle

%%%%%%%%%%%%%%%%%%%%%%%%%%%%%%%
\section{Introduction}
\label{sec:introduction}
%%%%%%%%%%%%%%%%%%%%%%%%%%%%%%%
Transparent data solutions is an emerging area of data management and analytics with a considerable impact on society. This is an important subject of debate in both engineering and law, involving scientists as well as activists and the press, because of the profound societal effects of such discrimination and biases.

In the last few years algorithms have been widely exploited in many practical use cases, thus they increasingly support and influence various aspects of our life. With little transparency in the sense that it is very difficult to ascertain why and how they produce a certain output, wrongdoing is possible. For example, algorithms can promote healthy habits by recommending activities that minimize risks only for a subset of the population because of biased training data. Whether these effects are intentional or not, they are increasingly difficult to spot due the opaque nature of machine learning and data mining.
Since algorithms affect us, transparent and better algorithms are indispensable  by making accessible not only the results of the data management and analysis but also the processes and models used. 

Today, the most efficient machine learning algorithms - such as deep neural networks - operate essentially as black boxes. Specifically, deep learning algorithms have an increasing impact on our everyday life: complex models obtained with deep neural network architectures represent the new state-of-the-art in many domains \cite{DBLP:journals/corr/Schmidhuber14} concerning image and video processing \cite{Krizhevsky:2012:ICD:2999134.2999257}, natural language processing \cite{Collobert:2008:UAN:1390156.1390177} and speech recognition \cite{6296526}.
However, neural network architectures present a natural propensity to opacity in terms of understanding data processing and prediction \cite{Strannegård2012,DBLP:journals/corr/abs-1710-00935}. This overall opacity leads to black-box systems where the user remains completely unaware of the process that models inputs over output predictions. Thus, with the introduction of complex, black-box systems, in the real world decision-making process, the need for algorithmic transparency becomes even more prominent.

This paper presents a new engine, named \XXX\ (\XXXdef ) to explain the main relationships between the inputs and outputs of a given prediction made by a black-box algorithm. As a first attempt \XXX\ explains predictions made by a convolutional neural network (CNN) \cite{726791} on image classification. To this aim, the main contributions of this work are threefold:
\begin{itemize}
\item 
Definition of a set of \textit{interpretable features} (input) characterizing the images through hypercolumn representation \cite{DBLP:journals/corr/HariharanAGM14a}. Hypercolums, representing pixels of a given image through all CNN layers, are clustered through K-means \cite{60082} to identify groups of correlated pixels. Each group models a given portion of the image representing an interpretable feature used to explain the black-box model.
\item
Definition of two indices to explain the behavior of the black-box model. The first, \IndexRI , measures the \textit{local influence of input feature with respect to the real class} of the image, while the second, \IndexB, measures the \textit{inter-class feature influence for each feature} of an image. Through these indices \XXX\ provides more insights on how a black-box model works.
\item
Definition of an iterative process of perturbation (based on blur) and classification to analyze the real impact/influence of a given interpretable feature over the local classification.
\end{itemize}
A preliminary experimental validation of \XXX\ performed on $85$ %\taniaTodo{\sout{aggiungere il numero delle immagini}} 
images demonstrate the effectiveness of \XXX\ in providing interesting relationships between a set of interpretable features characterizing the images and the class label selected through the CNN black-box model.

The paper is organized as follows. Section \ref{engine} provides a general overview of the \XXX engine providing process details in sections \ref{interpretable-feature-extraction} and \ref{influence-analysis}, while in section \ref{preliminary-results} some of the more interesting preliminary results are discussed with a detailed explanation of the meaning of the \IndexRI and \IndexB indexes.
Lastly, section \ref{discussion} provides a general discussion about other related works and some final considerations.
%%%%%%%%%%%%%%%%%%%%%%%%%%%%%%%
\section{The \XXX\ engine}
\label{engine}
%%%%%%%%%%%%%%%%%%%%%%%%%%%%%%%
\begin{figure}
\includegraphics[width=\linewidth]{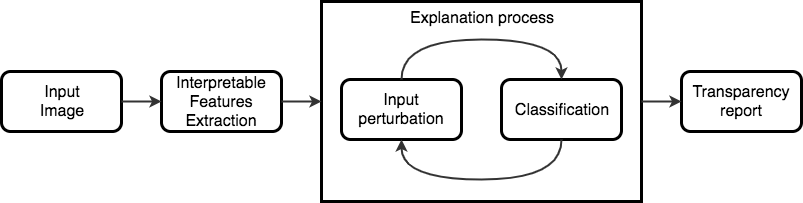}
\caption{Process}
\label{process-1}
\end{figure}
\XXX (\XXXdef) is a new data analytics engine to open up black-box algorithms by increasing their transparency. Its ultimate aim is to put existing, effective, and efficient algorithms to practical use cases. The \XXX\ engine explains the inner functioning of algorithms by providing explanations about the outcome produced through a deep convolutional neural network. \XXX\ analyzes the impact of each input feature on the final outcome (classification) through an iterative process based on input perturbation and classification and it has been tailored to the image processing and classification.

A convolutional network (CNN) \cite{726791} is a deep, feed-forward artificial neural network composed of many specialized hidden layers i.e. convolutional layers, pooling layers, fully connected layers and normalization layers. They have had great success in large-scale image and video recognition, achieving state-of-the-art accuracy on classification and localisation tasks, also thanks to very deep convolutional networks \cite{DBLP:journals/corr/SimonyanZ14a}. The main limitation of CNNs exploitation in many practical use cases is due to their opacity, i.e., their inner functioning is unclear. \XXX\ helps CNNs to be more transparent.

When given a pre-trained classification model, obtained through a black-box system (e.g., CNNs), \XXX\ identifies an interpretable explanation over a local classification. 
%\taniaTodo{Descrivere qui l'architettura con una breve spiegazione (1 frase) per ogni building block}
Figure \ref{process-1} shows the main building blocks of the \XXX\ architecture. If given an image, the \textit{Interpretable Features Extraction} step is performed through the hypercolumns extracted from the target convolutional model to identify a set of interpretable features (input) characterizing the images. An iterative perturbation of image features is then applied. At every iteration the system performs the classification on the perturbed image and produces a transparency report to provide details about how the algorithm made the prediction. To this aim, two innovative indices have been proposed.
The following sections describe the interpretable feature extraction process , and then address the generation of the transparency report.

\subsection{Interpretable feature extraction}
\label{interpretable-feature-extraction}
The first step towards the human-oriented analytics process is the definition of a set of interpretable features to correctly explain the forecasting/classification of a black-box model. The identification of this set of features when dealing with unstructured data, such as images and textual data, requires ad-hoc strategies to correctly ascertain why and how a given black-box classifier produces a certain output.  

Interpretable features should be neither too specific nor too general to effectively explain the classification outcome. In image processing, a single pixel of an image is both totally trifling and completely opaque in explaining how a black-box classifier produces a given output, whereas portions of image defined by a set of correlated pixels should be intuitively more effective.

To identify portions of image to be used as interpretable features \XXX\ performs a Simultaneous Detection and Segmentation (SDS) analysis \cite{DBLP:journals/corr/HariharanAGM14} based on hypercolumns \cite{DBLP:journals/corr/HariharanAGM14a} and cluster analysis via the K-Means algorithm \cite{60082}.

The SDS process is particularly suitable for this task because of its ability to segment the image in multiple portions, identifying the presence of multiple instances of the same object in an image. Figure \ref{pizza2-f1} shows a clear example of multiple instance identification with the $4$ highlighted items belonging to the same group of objects (in the specific case, $4$ pizzas).
Algorithms based on CNNs use the output of the last network layer to model the analyzed features. However, this layer usually produces a very coarse, not easily interpretable output, that cannot be used to explain the classification outcome. At the opposite side, earlier layers (hidden layers) are characterized by too many details, losing their semantic expressiveness. We believe that all the information contained in different CNN layers should be exploited to correctly explain the prediction outcome. Thus, hypercolumns provides an exhaustive behavioral description of the pixels through all the layers of the CNN.

Hypercolumns have been widely exploited in the SDS pipeline \cite{DBLP:journals/corr/HariharanAGM14a} yielding new state-of-the-art accuracy values in object detection and image segmentation. However in this work we use them with a different purpose and a slight variant in the implementation. In \XXX\ hypercolumns are used to identify correlated portions of the image instead of well defined objects, so the segmentation step is simpler (based on the K-means \cite{60082}) than the one described in \cite{DBLP:journals/corr/HariharanAGM14a}.

Given a black box CNN model composed of many layers and a labeled image belonging to a specific class, we compute the hypercolumns for each pixel of the image, as described in \cite{DBLP:journals/corr/HariharanAGM14a}. Specifically, given an image, we process it with a CNN and get only its representation through the most representative layers. A matrix of vectors is generated where each column in the matrix represents an input pixel through the relevant CNN layers.

\begin{figure}
\begin{subfigure}[b]{0.15\textwidth}
\includegraphics[width=\linewidth]{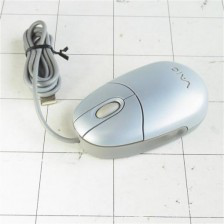}
\caption{}
\label{feature-extraction-original}
\end{subfigure}
\begin{subfigure}[b]{0.15\textwidth}
\includegraphics[width=\linewidth]{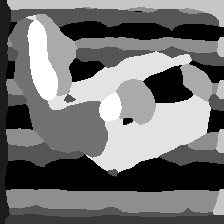}
\caption{}
\label{feature-extraction-clusters}
\end{subfigure}
\begin{subfigure}[b]{0.15\textwidth}
\includegraphics[width=\linewidth]{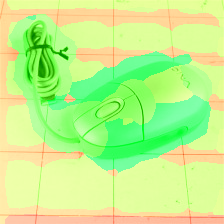}
\caption{}
\label{feature-report}
\end{subfigure}
\caption{Interpretable feature extraction example with $K = 10$. Figure \ref{feature-extraction-original} represents the original image. Figure \ref{feature-extraction-clusters} shows the image after the segmentation through hypercolumns clustering. Figure \ref{feature-report} shows the visual report produced by \XXX  }
\label{feature-extraction}
\end{figure}

Hypercolumns are then clustered exploiting the k-means algorithm to identify groups of pixels representing interpretable portions of the image with similar behavior through the most representative layers of the CNN model.  The output of the cluster analysis produces $k$ groups of correlated pixels corresponding to $k$ interpretable features. Figure \ref{feature-extraction} shows an example of interpretable feature extraction through hypercolumns and cluster analysis. It is noticeable how this strategy is able to identify homogeneous and highly interpretable portions of an image.

%%%%%%%%%%%%%%%%%%%%%%%%%%%%%%%
\subsection{Influence analysis}
\label{influence-analysis}
%%%%%%%%%%%%%%%%%%%%%%%%%%%%%%%
When given a set of interpretable features for a specific labeled image, \XXX\ performs an iterative process of input perturbation (based on blur) and classification to analyze the impact of input over the classifier output. First, \XXX exploits the black-box model to identify, for the original image, the set of probability values for each membership class. Then, for each interpretable feature, \XXX\ performs a blur perturbation of the original image in correspondence with a given feature and it uses the black-box model to predict the set of probability values for each membership class of the perturbed image.

Since our aim is to explain how the model works rather than assess how far the classification is accurate, we suppose that we know the label (membership class) of the original image. \XXX\ computes two indices to explain the black-box model behaviour:
\begin{itemize}
\item 
The \IndexRI\ index (\IndexRIDef) measures the \textit{local influence of the input feature with respect to the real class} of the image. 
\item
The \IndexB\ index (\IndexBDef)  measures the \textit{inter-class influence of input features} 
\end{itemize}

The \IndexRI\ index is calculated for each perturbed image as the ratio between the probability of belonging to the real class of the original image and the corresponding probability of the perturbed image. It ranges in $[0,\inf)$. 
When there are no features able to give the correct label to the image, \IndexRI\ is equal to $0$. On the other hand, there is no upper limit to the \IndexRI\ value. Specifically, \IndexRI\ assumes a very high value if the probability of belonging to the real class of the original image is very high and the probability of the perturbed version of the image has a value close to $0$. Both values are rare enough to be considered exceptions that never affect this kind of analysis.
In general \IndexRI\ values higher then $1$ represent a positive influence of the feature, while values lower then $1$ show a negative influence over the prediction of the real class.

To measure the influence of each input feature in the whole set of classes the \IndexB\ index is proposed. \IndexB\ is computed as the ratio between the \IndexRI\ value for the target class and the weighted average of \IndexRI\ for whole set of predicted classes, where the weights correspond to the probabilities of the predicted membership class for the original image.

\IndexB\ represents the ability of an input feature to uniquely represent the class of the original image. For \IndexB\ values lower than $1$ the input feature not only has an impact on the predicted class but also on all the others. Instead, if the \IndexB\ value is higher than $1$, the importance of the input feature for the real class is significant with respect to the whole set of predicted classes. Obviously, perturbations with \IndexB\ values close to $1$ can be considered neutral in the prediction process. 

\XXX\ produces a report as output for each image classification/prediction.
In the original image it highlights each feature's influence (figure \ref{feature-report}) and it provides details about the perturbation process along with the probability to belong to the real class and both \IndexRI\ and \IndexB\ values.

%%%%%%%%%%%%%%%%%%%%%%%%%%%%%%%
\section{Preliminary results}
\label{preliminary-results}
%%%%%%%%%%%%%%%%%%%%%%%%%%%%%%%
Some of the preliminary results obtained through  the \XXX\ system are discussed here below. 
\textit{Preliminary development and experimental settings.} \XXX\ is implemented in python and it exploits the features of Keras \cite{chollet2015keras}, a high-level neural network library, running on top of TensorFlow \cite{tensorflow2015-whitepaper}.
We exploit the K-means algorithm implemented in the scikit-learn python library \cite{scikit-learn} with the K-means++ initialization strategy.
The convolutional model selected for this preliminary work is the VGG-16 \cite{DBLP:journals/corr/SimonyanZ14a} developed by the VGG team from Oxford for the ImageNet competition ILSVRC-2014. It is a black-box model composed of $16$ layers (convolutional and fully connected layers) and it is able to predict, for each image, a membership class probability label from a predefined set of $1000$ classes. 
To identify the set of interpretable features for each image we consider the hypercolumns for the last $10$ layers of the CNN model. These layers correspond to the most representative ones.
Moreover, we experimentally define the $k$ parameter for the K-means to $10$.

\textit{Preliminary results} were obtained on a set of $85$ images of which $75$ belong to as many categories and $10$ belong to the same category \textit{pizza}.\\
Preliminary experiments address the evaluation of interpretable feature influence to explain block-box model.
Figure \ref{feature-extraction} shows the original version of a sample image belonging to class \textit{mouse}, the analysis of $10$ interpretable features and the visual report of feature influence produced by \XXX. The black-box model alone is not able to predict the \textit{mouse} class for the original image within the top 5 predictions as shown in Table \ref{mouse-original-predictions}. Indeed, the \textit{mouse} label is predicted just with $5.10\%$ of probability conversely to the wrong prediction of \textit{hand\_blower} with a probability of $26.20\%$ (see Table \ref{mouse-original-predictions}). \XXX\ explores the impact of each interpretable feature to understand what the reason behind the misleading prediction is. In Table \ref{mouse-feature-analysis} the analysis of $4$ interpretable features (i.e., 2 relevant/positive impact, 1 neutral and 1 irrelevant/negative impact features on the final prediction) of the mouse in Figure \ref{feature-extraction} is reported and analyzed. The first feature (row 1 in Table \ref{mouse-feature-analysis}) describes the contour of the mouse. When we perturbed this portion of the picture the probability of the image belonging to class of \textit{mouse} decreases. Therefore, the impact of the contours of the mouse has a positive impact on the prediction of the \textit{mouse} class and this is highlighted by the \IndexRI\ value equal to $5.10$. The third interpretable feature (row 3 in Table \ref{mouse-feature-analysis}) models top and right edges of the image and it can be considered neutral to the prediction. In fact the prediction of the \textit{mouse} class is slightly affected by the perturbation of this feature and this is reflected by the value of \IndexRI close to $1$.
The last interpretable feature reported in row 4 of Table \ref{mouse-feature-analysis} clearly highlights the line between floor tiles. By perturbing this feature, an increment for the \textit{mouse} class probability is obtained. Thus, the original model based its prediction on this feature. It is highlighted by the low \IndexRI\ value of $0.68$, suggesting that the black-box model mainly uses this feature to make the prediction.
Moreover, the \IndexB coefficient confirms the relevance, positive or negative, of each feature showing the same decreasing trend between \IndexB and \IndexRI.

\XXX\ summarizes the knowledge gained by \IndexRI\ through a visual report (Figure \ref{feature-report}). Each interpretable feature is colored according to the influence that it has on the prediction of the target class. Figure \ref{feature-report} is characterized by three different colors with different intensity: red describes a feature with a negative impact on the prediction of the target class, yellow represents a neutral feature for the class of the image and green shows a feature with a positive impact on the prediction of the correct class: the higher the intensity of the color, the higher the positive or negative influence.

\begin{table}[]
\centering
\begin{tabular}{|l|c|}
\hline
\textbf{Class}           & \multicolumn{1}{l|}{\textbf{P(C) \%}} \\ \hline
hand\_blower    & 26.20                        \\ \hline
washbasin       & 13.93                        \\ \hline
soap\_dispenser & 11.90                        \\ \hline
toilet\_seat    & 8.77                         \\ \hline
toilet\_tissue  & 7.35                         \\ \hline
\textbf{mouse}           & \textbf{5.10}                         \\ \hline
\end{tabular}
\caption{The top 5 predicted classes for the original image in Figure \ref{feature-extraction-original}. The real label of the image with the corresponding prediction is highlighted in bold.}
\label{mouse-original-predictions}
\end{table}

\begin{table}[]
\centering
\begin{tabular}{ccccc}
 \textbf{Features} & \textbf{Pertubations} & \textbf{P(c) \%} & \textbf{\IndexRI}   & \textbf{\IndexB}  \\ \hline
 \includegraphics[width=0.15\linewidth]{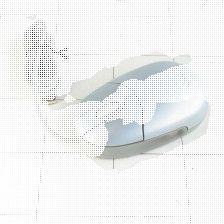}       & \includegraphics[width=0.15\linewidth]{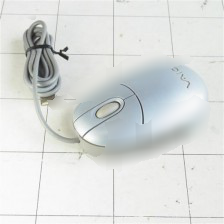}           & 0.99    & 5.10 & 2,96 \\ 
 \includegraphics[width=0.15\linewidth]{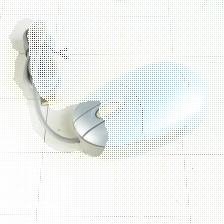}       & \includegraphics[width=0.15\linewidth]{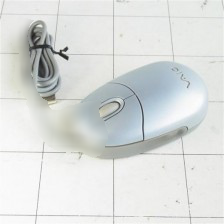}           & 1.17    & 2.87 & 2,00 \\ 
 \includegraphics[width=0.15\linewidth]{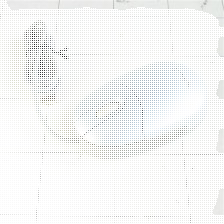}       & \includegraphics[width=0.15\linewidth]{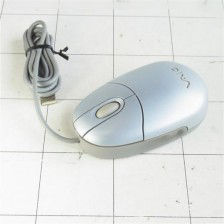}           & 4.67    & 1.09 & 1,09 \\ 
 \includegraphics[width=0.15\linewidth]{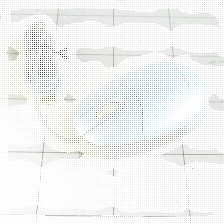}       & \includegraphics[width=0.15\linewidth]{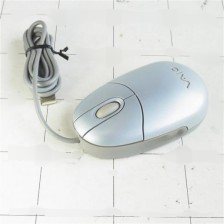}           & 7.49    & 0.68 & 0,36
\end{tabular}
\caption{Features perturbation impact evaluation.}
\label{mouse-feature-analysis}
\end{table}

\begin{figure}
\begin{subfigure}[b]{0.15\textwidth}
\includegraphics[width=\linewidth]{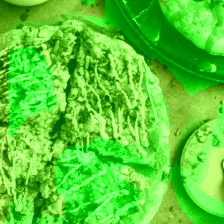}
\caption{}
\label{pizza0-report}
\end{subfigure}
\begin{subfigure}[b]{0.15\textwidth}
\includegraphics[width=\linewidth]{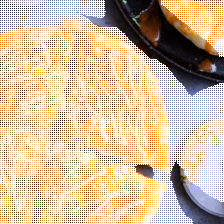}
\caption{}
\label{pizza0-f1}
\end{subfigure}

\begin{subfigure}[b]{0.15\textwidth}
\includegraphics[width=\linewidth]{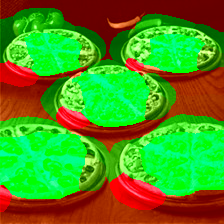}
\caption{}
\label{pizza2-report}
\end{subfigure}
\begin{subfigure}[b]{0.15\textwidth}
\includegraphics[width=\linewidth]{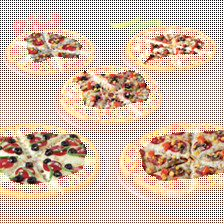}
\caption{}
\label{pizza2-f1}
\end{subfigure}
\caption{\IndexB evaluation}
\label{indexb-desc}
\end{figure}

\begin{figure}
\begin{subfigure}[b]{0.20\textwidth}
\includegraphics[width=\linewidth]{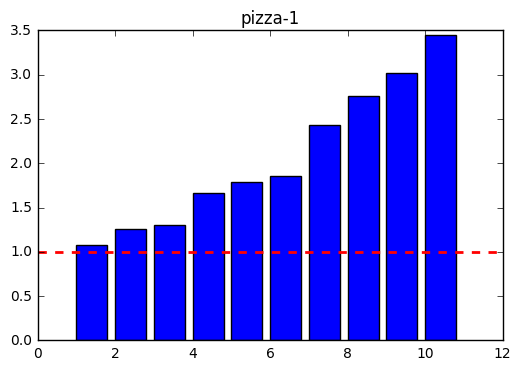}
\caption{}
\label{pizza-0-IR}
\end{subfigure}
\begin{subfigure}[b]{0.20\textwidth}
\includegraphics[width=\linewidth]{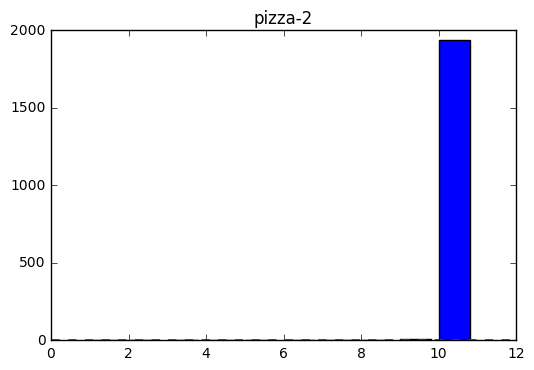}
\caption{}
\label{pizza-2-IR}
\end{subfigure}

\begin{subfigure}[b]{0.20\textwidth}
\includegraphics[width=\linewidth]{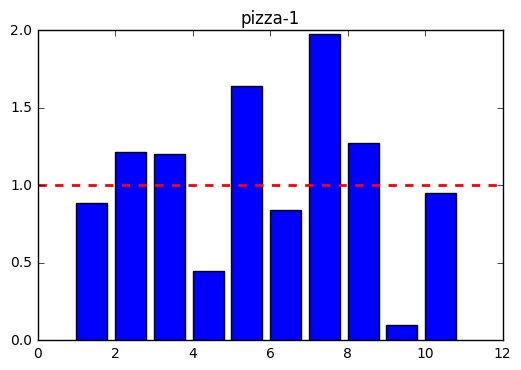}
\caption{}
\label{pizza-0-IRP}
\end{subfigure}
\begin{subfigure}[b]{0.20\textwidth}
\includegraphics[width=\linewidth]{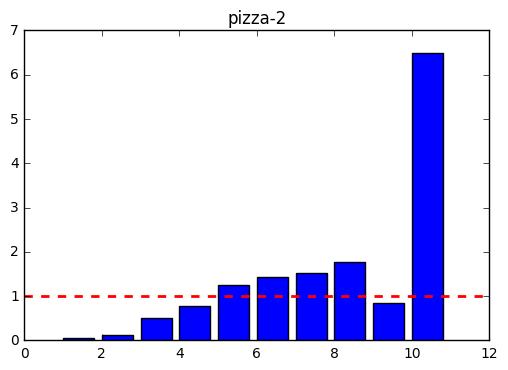}
\caption{}
\label{pizza-2-IRP}
\end{subfigure}
\caption{Relation between \IndexRI values (top) and \IndexB values (bottom) for the features of two images belonging to class \textit{pizza}.}
\end{figure}

Through \IndexRI\ and \IndexB\ \XXX\ makes it possible to distinguish between really useful features and misleading ones. Moreover, \XXX\ provides useful knowledge for understanding whether a feature with a positive or negative impact on the prediction uniquely identifies the target class with respect to the other predicted classes.
Figure \ref{indexb-desc} shows the report for two images belonging to the same class \textit{pizza} (Figures \ref{pizza0-report} and \ref{pizza2-report}) along with the selection of one of the most significant features analyzed by the model shown in Figure \ref{pizza0-f1} and Figure \ref{pizza2-f1} respectively.
For Figure \ref{pizza0-report} the model is not able to distinguish between the relevant features and the misleading ones with a predicted probability of belonging to class \textit{pizza} of $3.71\%$. Moreover, the visual report in Figure \ref{pizza0-report} shows a positive impact for all the extracted features. To understand the reason behind this behavior the \IndexB values for each feature should be analyzed. Figures \ref{pizza-0-IR} and \ref{pizza-0-IRP} show respectively the trend of the \IndexRI\ and the corresponding \IndexB value for each of 10 interpretable features. 
The feature in Figure \ref{pizza0-f1} corresponds to the second most positively influencing feature (9th bar in figure \ref{pizza-0-IR}) with $\IndexRI = 3.02$. However, the misleading knowledge contained in the feature is highlighted by the \IndexB bar chart shown in Figure \ref{pizza-0-IRP}. The \IndexB\ value for this feature is largely lower than $1$, with a value of $0.09$, meaning that this portion of the picture has a great influence not only on the target class but but also on a multitude of classes. Thus, in this case the model produces a wrong prediction because of the presence of features that positively influence many different classes. 
On the opposite side, the second report (Figure \ref{pizza2-report}) clearly shows how \XXX\ correctly understands the feature that positively influences the class \textit{pizza}. In this case there is a very influential feature (Figure \ref{pizza2-f1}) that has a positive effect that is noticeable in Figure \ref{pizza-2-IR}. Moreover, the \IndexB\ value of $6.49$ confirms the positive influence of this feature, represented by the last bar in Figure \ref{pizza-2-IRP}.

%%%%%%%%%%%%%%%%%%%%%%%%%%%%%%%
\section{Discussion}
\label{discussion}
%%%%%%%%%%%%%%%%%%%%%%%%%%%%%%%
The importance of algorithmic transparency and accountability is becoming even more relevant in our daily life \cite{DBLP:journals/corr/LepriSSLO16, DiakopoulosPaper}.
The quantity of data collected and analyzed with very complex algorithms in many different contexts is increasingly changing our lives. However, as algorithmic complexity increases, so the risk of misleading results increases as well: the more complex is the model, the more difficult it is to assert the reliability and fairness of the algorithmic decision-making process \cite{7546525}, thus also compromising the user's trust in the classification model even if outcomes are very accurate \cite{DBLP:journals/corr/RibeiroSG16}.

In the last few years some research efforts have been devoted to explaining the behavior of complex black-box models in different fields \cite{7546525,DBLP:journals/corr/RibeiroSG16,AlufaisanPaper,AdlerPaper}
and by presenting different metrics to evaluate the impact of input features on the final outcomes. The proposed techniques have been tailored to unstructured (e.g., image and text processing) \cite{DBLP:journals/corr/RibeiroSG16,DBLP:journals/corr/FongV17,DBLP:journals/corr/SimonyanVZ13,DBLP:journals/corr/abs-1710-00935} or structured data \cite{7546525}. Focusing on unstructured data, some works have put forward metrics for evaluating the impact of inputs on the classification outcomes \cite{DBLP:journals/corr/RibeiroSG16}, while others have exploited image segmentation \cite{DBLP:journals/corr/FongV17}, visualization methods \cite{DBLP:journals/corr/SimonyanVZ13}, or self-explaining techniques \cite{DBLP:journals/corr/abs-1710-00935}.

In this paper we have proposed \XXX, a new engine for black-box prediction-local explanation tailored to images. \XXX\ shows the explanation through visual reports and through the evaluations of two new indices: \IndexRI\ and \IndexB. Similarly to \cite{DBLP:journals/corr/RibeiroSG16} we analyzed the impact of a set of correlated pixels on the final classifier outcomes and we exploit a blur-perturbation approach as in \cite{DBLP:journals/corr/FongV17}. However we used a different technique based on hypercolumns representation jointly with cluster analysis to identify interpretable portions of correlated pixels exploiting the information contained in the black-box model, increasing the expressiveness of the explanation. Moreover, we introduce different indices to study the local influence of input features with respect to the real class of an image and the inter-class feature influence for each interpretable feature of an image. 
In particular, unlike other works \cite{DBLP:journals/corr/RibeiroSG16,DBLP:journals/corr/FongV17,7546525}, we take advantage of the architecture of the classification model to detect the real behavior of the algorithm, extracting an interpretable set of features that are significant and functional to the explanation of the classification.

This preliminary work opens the way to many possible future works such as the exploitation of local influence results to identify global influence explanations, the analysis of the explanation for different models, other than the extension of the \XXX\ system to the support of different types of unstructured data (e.g., text processing).

\bibliographystyle{ACM-Reference-Format}
\bibliography{EDBT2018}

\end{document}